\documentclass[runningheads]{llncs}
\usepackage{graphicx}
\usepackage{array}
\usepackage{booktabs}
\usepackage{longtable}
\usepackage{adjustbox}
\usepackage{tikz}
\usepackage{color}
\usepackage{times}

\newcommand*\circled[1]{\tikz[baseline=(char.base)]{
\node[shape=circle,draw,  inner sep=2pt, fill=blue] (char) {\bf{\textcolor{white}#1}};}}

\begin{document}
\title{No AI Without PI !}
\subtitle{Object-Centric Process Mining as the Enabler for Generative, Predictive, and Prescriptive Artificial Intelligence}
%
%
\author{Wil M.P. van der Aalst\inst{1,2}\orcidID{0000-0002-0955-6940}}
\authorrunning{Wil M.P. van der Aalst}
%
\institute{
Process and Data Science (Informatik 9), RWTH Aachen University, 52056 Aachen, Germany
\email{wvdaalst@pads.rwth-aachen.de} \and
Celonis, Theresienstrasse 6, 80333 Munich, Germany}
\maketitle              
\begin{abstract}
The uptake of Artificial Intelligence (AI) impacts the way we work, interact, do business, and conduct research.
However, organizations struggle to apply AI successfully in industrial settings where the focus is on end-to-end operational processes.
Here, we consider generative, predictive, and prescriptive AI and elaborate on the challenges of diagnosing and improving such processes.
We show that AI needs to be grounded using Object-Centric Process Mining (OCPM). Process-related data are structured and organization-specific
and, unlike text, processes are often highly dynamic.
OCPM is the missing link connecting data and processes and enables different forms of AI.
We use the term Process Intelligence (PI) to refer to the amalgamation of process-centric data-driven techniques able to deal with a variety of object and event types, enabling AI in an organizational context.
This paper explains why AI requires PI to improve operational processes and highlights opportunities 
for successfully combining OCPM and generative, predictive, and prescriptive AI.
\end{abstract}

\section{Introduction}
\label{sec:intro}

The gap between what we expect from today's information systems and reality is widening. 
Whereas organizations see the need to apply sophisticated \emph{Artificial Intelligence} (AI) techniques,
they struggle with rudimentary data management issues. 
Since the launch of OpenAI's ChatGPT in November 2022, \emph{Generative AI} (GenAI) has attracted a lot of attention and investments \cite{Dwivedi2023truncated}. 
Although many employees are using GenAI to improve their productivity (e.g. to produce presentations, reports, and programs),
processes are rarely impacted by GenAI. In its basic form, GenAI does not have access to process-related data and also more traditional forms of AI
need structured data to predict, diagnose, or circumvent process-related problems. 

Compare this to the ``smartphone paradox'', which refers to a striking contradiction in how we perceive the impact of new technologies 
and the actual real-world impact. On the one hand, there's a widespread feeling that the smartphone changed everything. 
It has transformed how we communicate, navigate, and consume information. Yet, when we step back and look more objectively,
especially at how work, organizations, and business processes have evolved,
the changes seem far more incremental than transformational. 
Most jobs still exist in similar forms as before, and organizations continue to rely on rather traditional information systems (e.g., ERPs, CRMs).
The smartphone has mostly optimized the edges of work rather than transforming the core of how work is structured and how products are created.
Another factor is that smartphones are more consumer-oriented than enterprise-oriented. 
The same principles apply to GenAI.

\emph{Process mining} has evolved from a research topic into a powerful approach for evidence-based process improvement supported by mature software tools. It was originally developed to bridge the gap between \emph{process science} and \emph{data science} 
by uncovering how work actually gets done in organizations \cite{process-mining-book-2016}. 
Process mining exposes bottlenecks, reveals deviations, and predicts problems before these happen.
Process mining is less visible than AI techniques that are used by individuals (e.g., OpenAI's ChatGPT, Google's NotebookLM, and Midjourney), 
just like traditional ERP systems like SAP and Oracle are not visible to most people, although all larger organizations depend on them.
Despite the lack of visibility of process mining, we argue that it is essential for the successful application of AI in organizations.
We aim to explain why recent breakthroughs in \emph{object-centric process mining} provide the ``grounding'' for \emph{generative}, \emph{predictive}, and \emph{prescriptive} AI.
\begin{figure}[thb]
\centerline{\includegraphics[width=12cm]{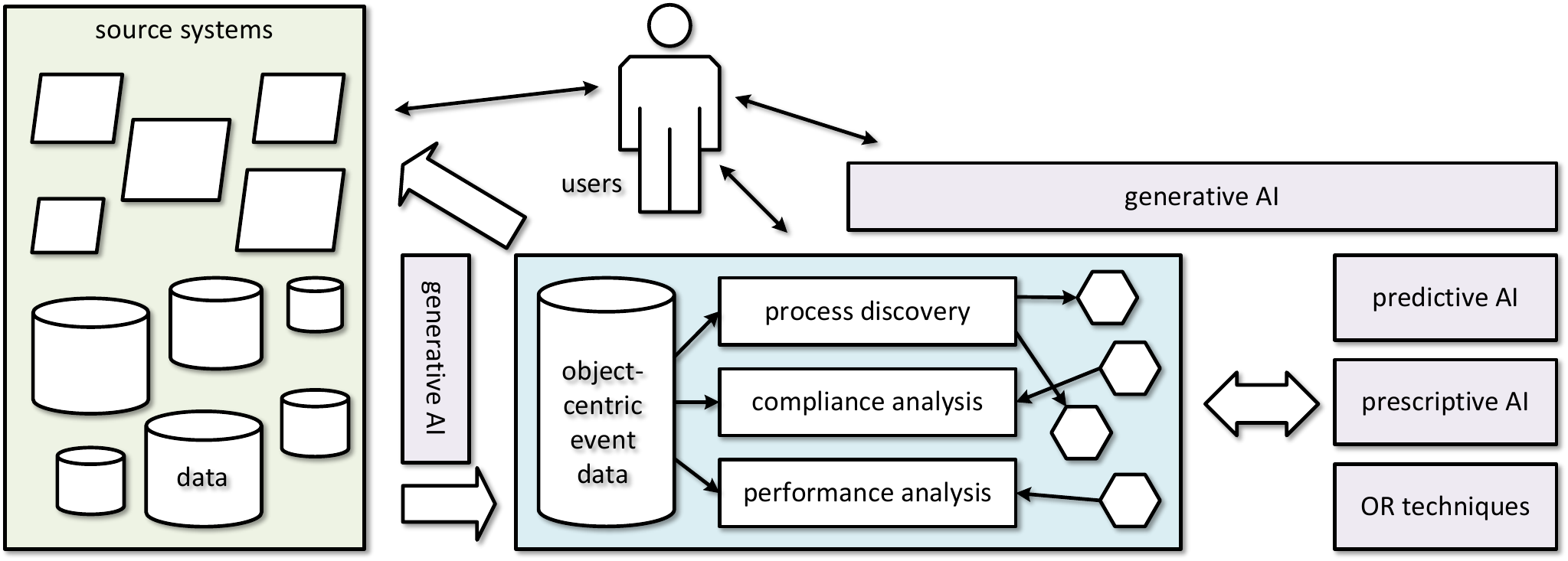}}
\caption{Overview showing how process mining capabilities (in the middle in blue) enable different types of AI to improve processes while leveraging existing source systems.}\label{fig-overview} 
\end{figure}

Figure~\ref{fig-overview} provides an overview of the connections between 
(1) the source systems used in organization, 
(2) object-centric process mining software, and 
(3) various forms of AI. 
\emph{Process Intelligence} (PI) refers to the combination of object-centric data-driven techniques enabling generative, predictive, and prescriptive AI in an organizational context \cite{process-mining-book-lars-2024,chapter-pm-book-lars-2024}.
This explains the provocative title ``No AI Without PI !'' stressing the importance of processes, process models, and process-related data.

The remainder of this paper is organized as follows.
Section~\ref{sec:ai} introduces the main forms of AI, followed by Section~\ref{sec:ocpm} 
which explains the importance and core ingredients of object-centric process mining.
Section~\ref{sec:aipi} discusses the relationships between both in detail, demonstrating that both complement each other. 
Section~\ref{sec:concl} concludes this keynote paper.

\section{Generative, Predictive, and Prescriptive Artificial Intelligence}
\label{sec:ai}

\emph{Artificial Intelligence} (AI) has been around for much longer than many realize. 
Its origins trace back to the 1950s, when pioneers like Alan Turing and John McCarthy laid its conceptual and technical foundations \cite{russell2016artificial}. 
Over the decades, AI evolved through cycles of optimism and skepticism.
Decades of seemingly incremental progress resulted in the tipping point we are currently experiencing.
The field of AI is broad, and there exist many classifications.
For much of its early history, AI was dominated by what's now called ``Good Old-Fashioned AI'' (GOFAI), i.e., 
logic-based systems that relied on symbolic reasoning and manually crafted rules.
With the rise of data-driven machine learning and neural networks \cite{goodfellow2016deep,NatureDL2015}, GOFAI moved to the background 
but still plays an important role (e.g., in automated planning). The same applies to \emph{Operations Research} (OR), which focuses on optimization, resource allocation, 
and mathematical modeling of decision problems (often under constraints and uncertainty) \cite{BENGIO2021405}.
\begin{table}[thb!]
\centering
\caption{Differences between generative, predictive, and prescriptive AI}\label{tab:ai}
\begin{adjustbox}{angle=90}
\begin{tabular}{|>{\raggedright}p{2.0cm}|>{\raggedright}p{3.7cm}|>{\raggedright}p{3.7cm}|>{\raggedright\arraybackslash}p{3.7cm}|}
\hline
\textbf{Criteria} & \textbf{Generative AI} & \textbf{Predictive AI} & \textbf{Prescriptive AI} \\
\hline
\textbf{Core Function} & Generates new data (e.g., text, images, or code) resembling the training data but guided by a prompt. & Predicts future outcomes using models trained on past data and applied to new data. & Suggests or enforces actions based on models and predefined goals. \\
\hline
\textbf{Goal} & To create new content or data that mimics reality. & To forecast future events or behaviors from historical data. & To recommend decisions or actions that optimize key performance metrics. \\
\hline
\textbf{Methodology} & Uses generative models to produce new data using a random seed or prompt. & Utilizes statistical tools, mining techniques, and machine learning models to predict outcomes. & Applies predictive models or optimization algorithms constrained by rules and driven by goals. \\
\hline
\textbf{Application Examples} & Automated text and image generation. Summarization of documents. Code and query generation. & Risk assessment, congestion forecast, churn prediction. & Resource allocation, real-time decision making, automated planning.\\
\hline
\textbf{Outcome Type} & Creative or mimetic content that seems generated by humans. & Statistical probabilities or predicted outcomes regarding future events. & Actions to address forecasted scenarios or automated decisions. \\
\hline
\textbf{Data Needs} & Requires huge amounts of diverse data to learn content generation. Data is typically general and not problem/situation specific. & Needs historical data relevant to the specific outcomes being predicted, i.e., data need to be specific for the task at hand. & Uses models constrained by rules and driven by goals, i.e., model-driven. \\
\hline
\textbf{Key Benefits} & Enhances creativity, provides data augmentation, and mimic human behavior & Helps to anticipate trends, problems, and opportunities. & Enhances operational efficiency and strategic decision-making by optimizing actions. \\
\hline
\end{tabular}
\end{adjustbox}
\end{table}

Table~\ref{tab:ai} lists the key characteristics of \emph{generative}, \emph{predictive}, and \emph{prescriptive} Artificial Intelligence (AI).
The boundaries between the different types of AI are not crisp. AI solutions often combine many ``ingredients'' and there is a huge diversity of approaches. 

\emph{Generative AI} (GenAI) can be best understood by looking at the problem of predicting the next word. Assume one counts the frequency of $n$-grams in the seven Harry Potter books of Joanne Rowling. For example, for $n=4$ one finds 4-grams like $\sigma_1 = \allowbreak \langle$ ``turned'',  ``on'', ``his'', ``heel'' $\rangle$ and $\sigma_2 = \allowbreak \langle$ ``defence'',  ``against'', ``the'', ``dark'' $\rangle$ happening frequently.
By comparing these frequencies with the frequencies of 3-grams like $\sigma_3 = \allowbreak \langle$ ``turned'',  ``on'', ``his'' $\rangle$ and $\sigma_4 = \allowbreak \langle$ ``defence'',  ``against'', ``the'' $\rangle$, one can create conditional probabilities for the next word. 
For example, if  $\mathit{freq}(\sigma_1) = 20$ and  $\mathit{freq}(\sigma_3) = 100$, then the probability that the text fragment ``turned on his'' is followed by the word ``wheel'' is 0.2. Given a small prefix (sometimes called ``prompt''), one can recursively generate the next word using these probabilities.
If the word ``wheel'' is generated after ``turned on his'', then the prefix ``on his wheel'' can be considered to predict the next word. Etc.
This is a random process, i.e., different texts can be created using a different initial seed.
Doing this for the seven Harry Potter books results in texts that seem to be written by Joanne Rowling.
Another example illustrating generative AI is the application of \emph{Generative Adversarial Networks} (GANs) \cite{goodfellow2014generative}
which consist of pairs of neural networks:
a generator that creates fake data and a discriminator that tries to tell real from fake.
Both compete in a game, trying to improve their performance over time. 
As the generator learns to fool the discriminator, it gets better at producing data that closely mimics the real thing, 
whether it's images, audio, or text.
A \emph{foundation model} is a large, general-purpose model trained on broad data (often at scale) that can be adapted or fine-tuned for many downstream tasks -- like translation, summarization, question answering, etc.
Examples of such models focusing on textual data, also called \emph{Large Language Models} (LLMs), are 
ChatGPT (OpenAI), Claude (Anthropic), and Gemini (Google).

\emph{Predictive AI} can be viewed as trying to learn a function $f \in X \rightarrow Y$ based on many examples.
$X$ is the domain of the function describing the descriptive features and $Y$ is the range describing one or more target features.
Features can be categorical or numerical. Consider, for example, the problem of predicting someone's yearly income based on gender $x_1$, age $x_2$, and education level $x_3$. $f(x_1,x_2,x_3)$ is the predicted income based on the three descriptive features.
One can train a model based on many examples of the form $((x_1,x_2,x_3),y)$. For example, $((\mbox{female},\mbox{56},\mbox{PhD}),\mbox{150k})$.
Learning $f$ corresponds to trying to minimize the error $|f(x_1,x_2,x_3)-y|$ over all training examples.
Note that the function may be represented by a simple linear function or a complex neural network \cite{bishop2006pattern,goodfellow2016deep,mitchellbook}.

\emph{Prescriptive AI} often involves a goal and constraints. 
One can think of this as function $f \in X \rightarrow Y$, 
however, now the goal is not to predict a target feature, but to pick the best possible outcome.
Consider the following descriptive features in a credit rating application: 
gender $x_1$, age $x_2$, education level $x_3$, and yearly income $x_4$.
Based on this we may need to decide whether the person gets a credit, i.e., $Y = \{ \mbox{Credit},\mbox{NoCredit}\}$.
For example, if $f(\mbox{female},\mbox{56},\mbox{PhD},\mbox{150k}) = \mbox{Credit}$, then the person gets the credit.
Function $f$ tries to optimize a predefined goal, e.g., 
maximize the number of customers paying their debts minus the number of customers not paying their debts.
The ``optimal'' function may be learned based on data. However, it may also be the case that classical mathematical optimization is used, 
e.g., $f$ corresponds to a Mixed-Integer Linear Program (MILP).

The above descriptions are rather simplistic and are only used to characterize the three types of AI discussed in the context of process intelligence.
Especially with the uptake of LLMs, it is not so clear how AI applications work precisely.
For example, ChatGPT utilizes Python in the background to perform computations and uses the internet to look up information (e.g., current prices or weather). The question ``What is the distance from Amsterdam to Rome to Brisbane?'' is answered correctly by looking up the three locations and adding up the two distances computed using the haversine formula (i.e., the shortest distance between two points on the surface of a sphere).

\emph{Retrieval-Augmented Generation} (RAG) is a crucial extension for utilizing GenAI in an enterprise setting.
LLMs do not have access to real-time data (prices, weather), very specific data (e.g., the precise location of a city), 
and private data (e.g., sales orders). LLMs are created using public, stable, and general-purpose information.
RAG solves this by grounding the model's responses in retrieved facts \cite{lewis2020retrieval}.
Therefore, it combines two mechanisms: 
(1) \emph{retrieval}, i.e., pulling in relevant information from an external source (like a database, document collection, or even the web), and
(2) \emph{generation}, i.e., generating a natural-language response based on that information or posing new retrieval questions.
In this way, GenAI applications enhance the generation process with contextually rich, pre-existing or real-time information, 
leading to more informed and accurate outputs.

\section{Object-Centric Process Mining}
\label{sec:ocpm}

Traditional \emph{case-centric process mining} involves analyzing event logs to extract process models, 
check conformance, analyze performance and support process improvements \cite{process-mining-book-2016}. 
It focuses on sequences of activities related to cases, such as a specific order, claim, patient, or application. 
Process models such as Petri nets, Directly-Follows-Graphs (DFGs), and BPMN models play an important role and serve as grounding for event data.
This approach is distinct from the AI approaches presented in the previous section.
While AI encompasses a wide range of data-driven methods for pattern recognition and prediction, 
process models play no role, and mainstream AI techniques cannot be used to discover, control, and improve \emph{end-to-end} processes.
If used in an enterprise setting, AI is often used to automate or manage a single task.
Process mining reveals the underlying control-flow of processes, uncovers bottlenecks, and checks for compliance in a way that is interpretable and actionable for domain experts.
Process mining has seen rapid adoption across industries such as manufacturing, finance, healthcare, and the public sector. The increasing demand for transparency, efficiency, and digital transformation has fueled this growth. There are over 50 vendors of process mining software, and 
Gartner, and other analyst firms have identified process mining as a new and important category of tools \cite{gartner-MQ-PM-2024}.

However, organizations adopting traditional process mining have identified several recurring challenges \cite{mathematics-OCPM-wvda-2023}:

\begin{itemize}
\item \emph{Scattered data}: Information lives across dozens of systems and tables, making data extraction a time-consuming task.
Finding the data and transforming it into event logs may be a time-consuming task.
\item \emph{Rigid case notion}: Traditional tools assume every process can be captured in a single ``case'',  which results in interconnected partly overlapping process models. Issues such as delays or compliance failures often stem from interactions between departments or processes.
\item \emph{Oversimplified distorted views}: Events often involve multiple objects, but forcing them into a single-case model distorts the truth.
\item \emph{Inactionable insights}: Many tools provide post-mortem analysis, not real-time, actionable diagnostics.
It is important to continuously load event data and provide techniques that turn insights into actions.
This requires a connection to automation and AI tools.
\item \emph{Cultural resistance}: Even with the right tools, change is hard. 
Process mining makes operational problems and undesirable behaviors transparent. This often creates resistance.  
Without the buy-in of top-level management and clear communication, organizations struggle to implement the identified improvement opportunities.
\end{itemize}

Apart from the last challenge, which is more of an organizational nature, these challenges can be addressed by a combination
of \emph{Object-Centric Process Mining }(OCPM) and AI leading to what we call \emph{Process Intelligence} (PI).
\begin{figure}[thb]
\centerline{\includegraphics[width=12cm]{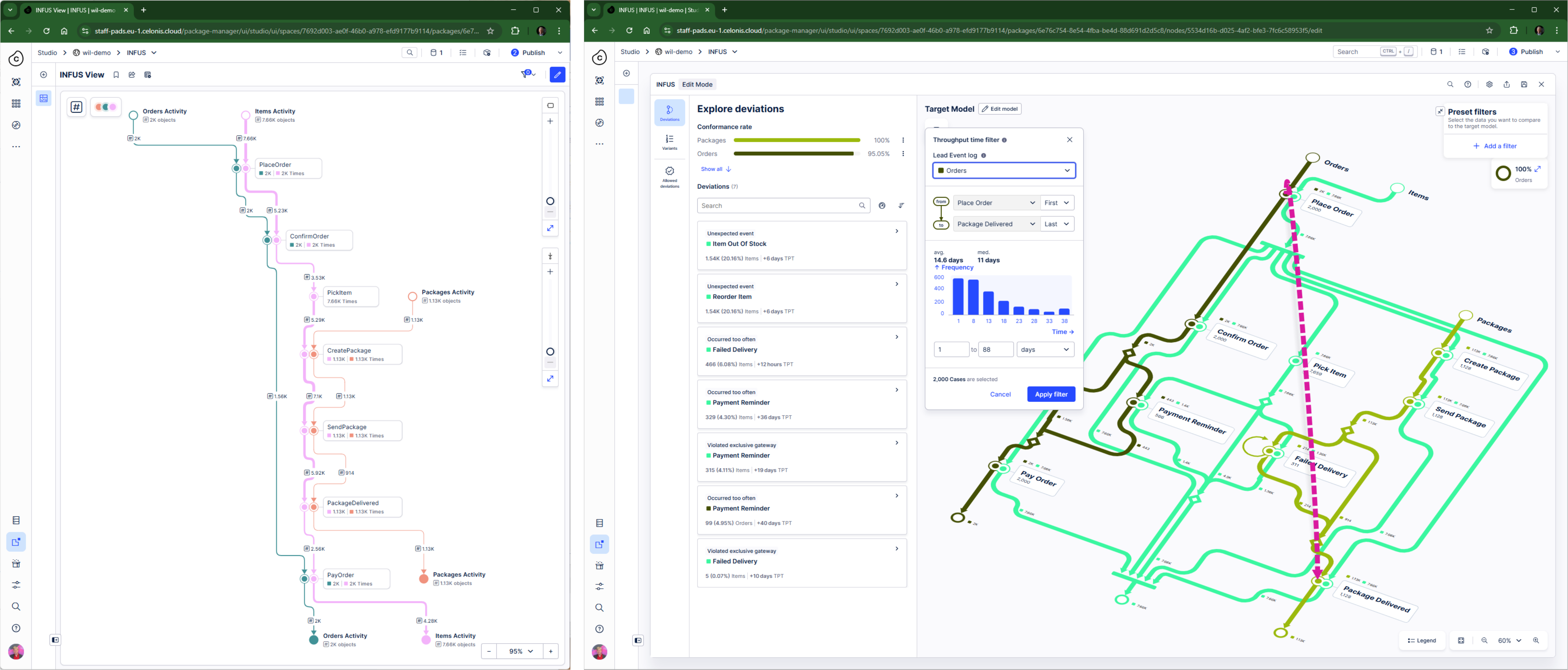}}
\caption{Two screenshots illustrating OCPM in Celonis. The left-hand side shows a discovered object-centric DFG using three object types: orders, items, and packages. The right-hand side shows an object-centric BPMN model using the same three object types, showing a list with the most frequent compliance problems, and a throughput time analysis showing the time between placing an order and delivering the last package containing items from that order.}\label{fig-ocpm} 
\end{figure}

OCPM starts from \emph{Object-Centric Event Data} (OCED) where events can reference multiple objects of different types, 
such as a payment linked to both an invoice and a customer, or a production order linked to different parts.
Recall that, in traditional case-centric process mining, a case refers to precisely one case. This is natural when considering 
process modeling notations such as flowcharts, BPMN models, DFGs, workflow nets, etc.\ that use the same assumption.
However, in reality, case-centricity is very limiting. Using OCPM it is possible to discover process models describing different types of objects in a single model \cite{ocpn_fi_2020}. Moreover, techniques for conformance checking and performance analysis have been extended to this more general setting. Figure~\ref{fig-ocpm} shows the Celonis OCPM software in action. We refer to \cite{ocpm-white-paper-celonis-2023,mathematics-OCPM-wvda-2023} for more details.

By adopting OCPM, organizations gain the ability to analyze operational activities from any perspective using a unified, consistent dataset, i.e., a \emph{single source of truth}. Instead of relying on system-specific event logs, 
event data should ideally be system-agnostic. For example, a process executed in SAP should generate the same kind of event data as that same process in Oracle. Storing data as OCED, e.g., using the OCEL 2.0 format \cite{OCELpage2024}, makes this possible.

With OCPM, there is no need to re-extract or reshape data when shifting the analytical viewpoint. This enables flexible, on-demand process mining views tailored to different questions or stakeholders. OCPM uncovers valuable insights, especially for issues that occur at the intersections of different processes and organizational units, i.e., areas that are often overlooked with traditional methods.

\section{``No AI Without PI''}
\label{sec:aipi}

Organizations struggle to leverage the amazing advances in AI  (cf.\ Section~\ref{sec:ai}) to enhance end-to-end processes. 
It is possible to automate or accelerate selected tasks, but it is unclear how to apply AI \emph{at the enterprise level}.
However, OCPM provides the missing \emph{grounding} for AI by exposing operational processes and their data.
\begin{figure}[thb]
\centerline{\includegraphics[width=9cm]{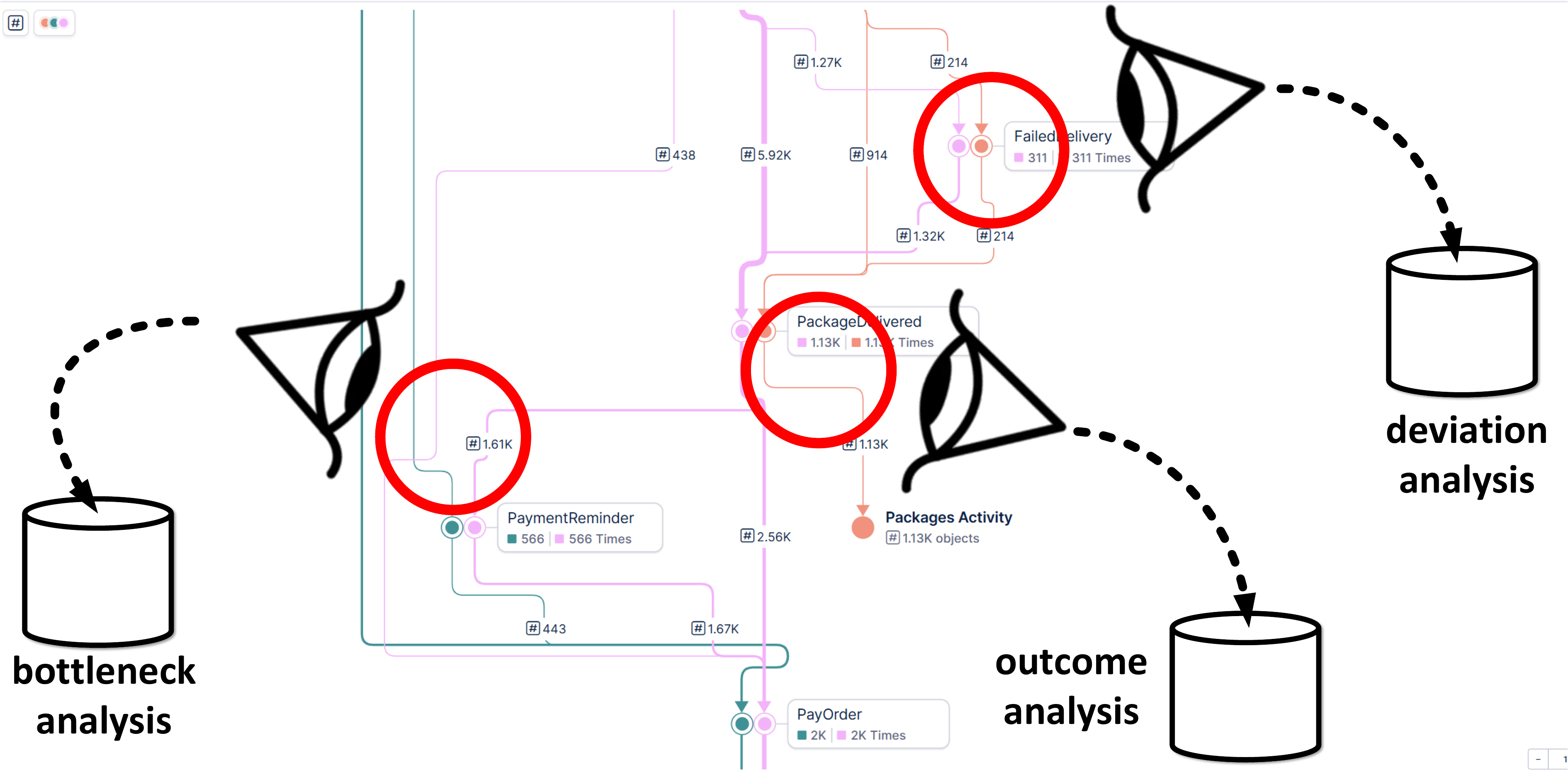}}
\caption{OCPM serves as the lens to apply AI in an enterprise setting. 
Any process-related problem (e.g., a bottleneck) can be translated into a machine learning problem for diagnosis and prediction.}\label{fig-eye} 
\end{figure}

Figure~\ref{fig-eye} illustrates how OCPM helps to create the \emph{context} required for AI. 
Without process mining, it is impossible to talk about process-related problems such as bottlenecks and deviations.
However, after discovering process models and connecting these to the actual data, it is possible to create, 
for example, machine learning problems.
\begin{figure}[thb]
\centerline{\includegraphics[width=10cm]{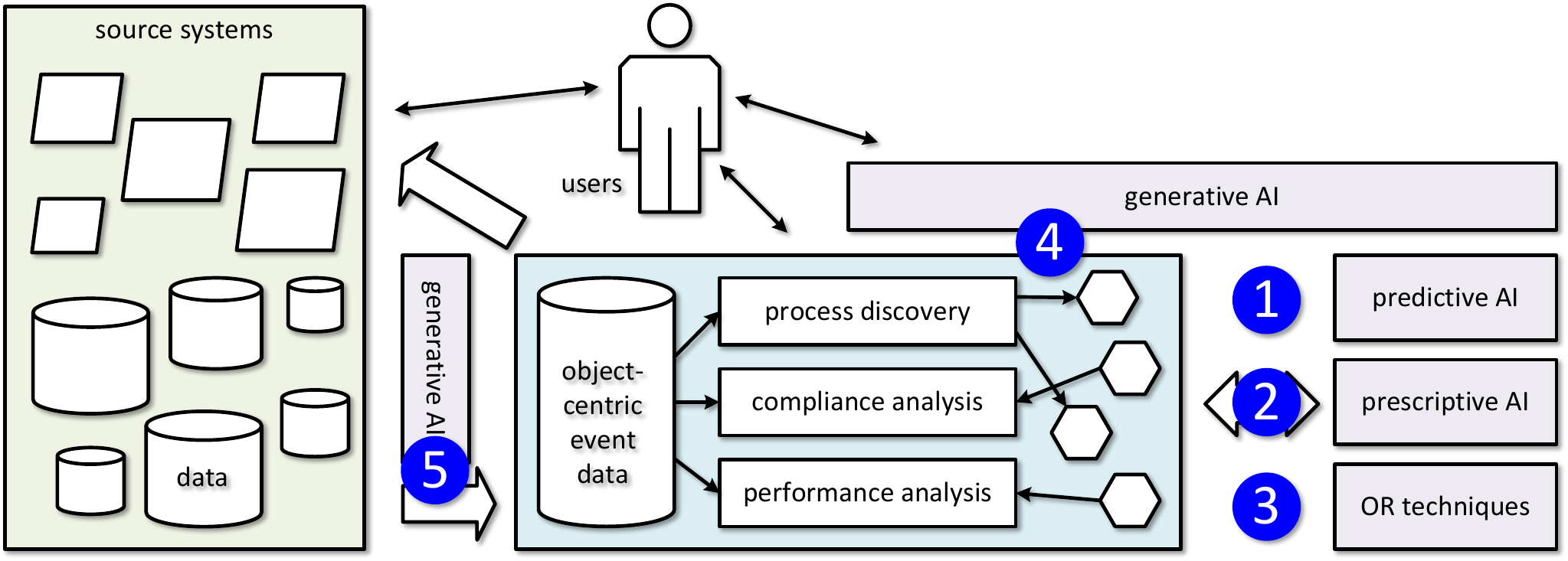}}
\caption{Based on Figure~\ref{fig-overview}, we highlight five opportunities to combine process mining and AI.}\label{fig-refined} 
\end{figure}

Figure~\ref{fig-refined} shows five connections \circled{1}--\circled{5} between OCPM as presented in Section~\ref{sec:ocpm} and the different forms of AI discussed in Section~\ref{sec:ai}. As explained, predictive AI is about learning a function from examples.
For any process related problem, we can create training examples \circled{1} and derive the corresponding function using machine learning.
For example, given a bottleneck, we can create a training instance each time we visit the bottleneck and use the experienced delay as the target feature.
Figure~\ref{fig-eye} illustrates the basic idea. Note that the data generated for each process-related problem is often tabular with a selected target feature. However, this does not need to be the case. Fragments of OCED may also be represented as graph and there can be multiple target features.
A lot of work has been done on this, typically using terms such as \emph{operational support} \cite{process-mining-book-2016} and \emph{predictive process monitoring}
\cite{DiFrancescomarino2022}. Prescriptive AI \circled{2} is connected in a similar way to the process-mining engine. However, the focus is not on taking actions or making decisions. Also, models are goal-driven and respect predefined constraints.
As mentioned, the boundaries between prescriptive AI and Operations Research (OR) are not crisp. 
Classical optimization, scheduling, and planning techniques \circled{3} can be integrated in the same way.
Also note that some process mining techniques (e.g., computing alignments) use mathematical optimization inside \cite{process-mining-book-2016}.
Just like machine learning can guide optimization \cite{BENGIO2021405}, machine learning can guide process discovery \cite{Vincenzo-InfSci2022}.

The two remaining connections refer to the use of generative AI (GenAI). GenAI can be used to make interactions between process mining software and humans easier \circled{3}. Users can pose questions in natural language and the engine can describe the OCED (e.g., which objects, events, and relations exist) and process mining capabilities (e.g., performance indicator functions and analysis techniques). A GenAI like ChatGPT can combine both and then use the Retrieval-Augmented Generation (RAG) approach described in Section~\ref{sec:ai}. This ensures that answers are based on process mining computations rather than guessing based on a general-purpose LLM.
In \cite{alessandro-bpm/Berti0A23} we show that just sending textually encoded process variants or DFGs to the GenAI is enough to generate answers, but these are not very reliable. Yet, the interplay between domain knowledge, process discovery, and LLMs may provide novel insights. GenAI may also be used to assist in creating normative models \cite{humam-llm-mod-BPMDS-EMMSAD-2024}.
The benchmark presented in \cite{PM-LLM-Benchmark-arXiv-2024} provides an extensive set of repeatable process-related challenges distributed over seven categories. All new LLMs are evaluated to see how process-analysis capabilities improve over time. Currently (March 2025), close to 100 LLMs have been evaluated and ``o3-mini-20250131-HIGH'' ranks highest. Another valuable use case for using AI in the context of process mining is the preparation of event data \circled{5}. Although this is a considerable bottleneck in the adoption of process mining, research on this topic is limited.
OCPM makes data extraction easier because OCED allows for data representations closer to reality, and there is no need to repeatedly revisit the source systems when questions change. However, there is still a need to bridge the gap between the specifics of proprietary data formats and OCED. GenAI can help to facilitate this (see for example the high-quality SQL generation by most LLMs).

\section{Conclusion}
\label{sec:concl}

\emph{It is the process, stupid!} This is a phrase frequently used in the field of process management.
Often, we only realize there is a process when something does not work out as planned, e.g., 
a flight is delayed, a request is never answered, an order is lost, etc.
Broken processes lead to broken promises, waste, delays, rework, frustration, and financial losses.
AI has the potential to ensure that processes work, organizations thrive, innovation accelerates, 
and even the planet benefits (e.g., reduced emissions). However, the ``smartphone paradox'' (Section~\ref{sec:intro}) 
reveals that we sometimes overestimate the impact of new technologies in an enterprise setting.
AI already takes care of selected tasks but is rarely used to improve end-to-end processes.

Therefore, we elaborated on the relationship between process mining and AI.
We characterized the different forms of AI and the connection to Object-Centric Process Mining (OCPM).
Figure~\ref{fig-refined} showed five connections between AI and OCPM.
AI needs to be grounded in the organization’s processes and related data.
OCPM provides the grounding needed. Without this context, AI applications will be isolated.
Therefore, we advocated Process Intelligence (PI), i.e., the combination of both.
\emph{When processes work, everything works!}

\bibliographystyle{plain}
\bibliography{lit}

\end{document}